\documentclass[11pt]{article}
\usepackage[latin9]{inputenc}
\usepackage{array}
\usepackage{booktabs}
\usepackage{amsmath}
\usepackage{graphicx}

\makeatletter

\providecommand{\tabularnewline}{\\}

\@ifundefined{date}{}{\date{}}
\usepackage{acl2016_latest}
\usepackage{times}
\usepackage{latexsym}

\aclfinalcopy % Uncomment this line for the final submission
\title{A Compositional Approach to Language Modeling}

\author{Kushal Arora\\
    Department of CISE, \\
    University of Florida\\
    {\tt karora@cise.ufl.edu}
  \And
Anand Rangarajan\\
  	Department of CISE\\
  	University of Florida\\
  {\tt anand@cise.ufl.edu}}

\date{}

\makeatother

\begin{document}
\maketitle
\begin{abstract}
Traditional language models treat language as a finite state automaton
on a probability space over words. This is a very strong assumption
when modeling something inherently complex such as language. In this
paper, we challenge this by showing how the linear chain assumption
inherent in previous work can be translated into a sequential composition
tree. We then propose a new model that marginalizes over all possible
composition trees thereby removing any underlying structural assumptions.
As the partition function of this new model is intractable, we use
a recently proposed sentence level evaluation metric Contrastive Entropy
to evaluate our model. Given this new evaluation metric, we report
more than 100\% improvement across distortion levels over current
state of the art recurrent neural network based language models.
\end{abstract}

\section{Introduction\label{sec:Introduction}}

The objective of language modeling is to build a probability distribution
over sequences of words. The traditional approaches, inspired by Shannon's
game, has molded this problem into merely predicting the next word
given the context. This leads to a linear chain model on words. In
its simplest formulation, these conditional probabilities are estimated
using frequency tables of word $w_{i}$ following the sequence $w_{i-1}^{1}$.
There are two big issues with this formulation. First, the number
of parameters rises exponentially with the size of the context. Second,
it is impossible to see all such combinations in the training set,
however large it may be. In traditional models, the first problem,
famously called the \emph{curse of dimensionality,} is tackled by
limiting the history to the previous $n-1$ words leading to an n-gram
model. The second problem, one of sparsity, is tackled by redistributing
the probability mass over seen and unseen examples usually by applying
some kind of smoothing or interpolation techniques. A good overview
of various smoothing techniques and their relative performance on
language modeling tasks can be found in \cite{goodman2001bit}.

Smoothened n-gram language models fail on two counts: their inability
to generalize and their failure to capture the longer context dependencies.
The first one is due to the discrete nature of the problem and the
lack of any kind of implicit measure of relatedness or context based
clustering among words and phrases. The second problem---the failure
to capture longer context dependencies---is due to the n-order Markov
restriction applied to deal with the curse of dimensionality. There
have been numerous attempts to address both the issues in the traditional
n-gram framework. Class based models \cite{brown1992class,baker1998distributional,pereira1993distributional}
try to solve the generalization issue by deterministically or probabilistically
mapping words to one or multiple classes based on manually designed
or probabilistic criteria. The issue of longer context dependencies
has been addressed using various approximations such as cache models
\cite{kuhn1990cache}, trigger models \cite{lau1993trigger} and structured
language models \cite{charniak2001immediate,chelba1997structure,chelba2000structured}.

Neural network based language models take an entirely different approach
to solving the generalization problem. Instead of trying to solve
the difficult task of modeling the probability distribution over discrete
sets of words, they try to embed these words into a continuous space
and then build a smooth probability distribution over it. Feedforward
neural network based models \cite{bengio2006neural,mnih2009scalable,morin2005hierarchical}
embed the concatenated n-gram history in this latent space and then
use a \emph{softmax} layer over these embeddings to predict the next
word. This solves the generalization issue by building a smoothly
varying probability distribution but is still unable to capture longer
dependencies beyond the Markovian boundary. Recurrent neural network
based models \cite{mikolov2011extensions,mikolov2010recurrent} attempt
to address this by recursively embedding history in the latent space,
predicting the next word based on it and then updating the history
with the word. Theoretically, this means that the entire history can
now be used to predict the next word, hence, the network has the ability
to capture longer context dependencies.

All the models discussed above solve the two aforementioned issues
to varying degrees of success but none of them actually challenge
the underlying linear chain model assumption. Language is recursive
in nature, and this along with the underlying compositional structure
should play an important role in modeling language. The computational
linguistics community has been working for years on formalizing the
underlying structure of language in the form of grammars. A step in
the right direction would be to look beyond simple frequency estimation-based
methods and to use these compositional frameworks to assign probability
to words and sentences.

We start by looking at n-gram models and show how they have an implicit
sequential tree assumption. This brings us to the following questions:
Is a sequential tree the best compositional structure to model language?
If not, then, what is the best compositional structure? Further, do
we even need to find one such structure, or can we marginalize over
all structures to remove any underlying structural assumptions?

In this paper we take the latter approach. We model the probability
of a sentence as \textbf{\emph{the marginalized joint probability
of words and composition trees}} (over all possible rooted trees).
We use a probabilistic context free grammar (PCFG) to generate these
trees and build a probability distribution on them. As generalization
is still an issue, we use distributed representations of words and
phrases and build a probability distribution on them in the latent
space. A similar approach but in a different setting has been attempted
in \cite{socher2013parsing} for language parsing. The major difference
between our approach and theirs is the way we handle the breaking
of PCFG's independence assumption due to the distributed representation.
Instead of approximating the marginalization using the n-best trees
as in \cite{socher2013parsing}, we restore this independence assumption
by averaging over phrasal representations leading to a single phrase
representation. This single representation for phrases in turn allows
us to use an efficient Inside-Outside \cite{lari1990estimation} algorithm
for exact marginalization and training.

\section{Compositional View of an N-gram model\label{sec:Compositional-n-gram}}

\label{gen_inst}

Let us consider a sequence of words $w_{1}^{n}$. A linear chain model
would factorize the probability of this sentence $p(w_{1}^{n})$ as
a product of conditional probabilities $p(w_{i}|h_{i})$ leading to
the following factorization:

\begin{equation}
p(w_{1}^{n})=\prod_{i=1}^{n}p(w_{i}|h_{i}).
\end{equation}

\begin{figure}[t]
\begin{centering}
\includegraphics[clip,scale=0.8]{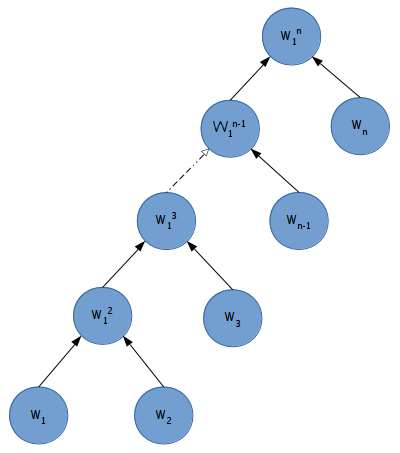}
\par\end{centering}

\caption{Sequential tree of a linear chain model\label{fig:seq-tree-n-gram}}
\end{figure}
All the models discussed in previous section differ in how the history
or context $h_{i}$ is represented. For a linear chain model with
no assumptions, the history $h_{i}$ would be the previous $i-1$
words, so the factorization is
\begin{equation}
p(w_{1}^{n})=\prod_{i=1}^{n}p(w_{i}|w_{1}^{i-1}).\label{eq:n-gram-pure}
\end{equation}

If we move the probability space to sequences of words, the same factorization
in equation~(\ref{eq:n-gram-pure}) can be written as:
\begin{multline}
p(w_{1}^{n},\ldots,w_{1}^{i},\ldots,w_{1}^{2},w_{n},\ldots,w_{i},\ldots,w_{1})=\\
\prod_{i=1}^{n}p(w_{1}^{i}|w_{1}^{i-1},w_{i})p(w_{i}).\label{eq:seq-factorization}
\end{multline}
Figure~\ref{fig:seq-tree-n-gram} shows the sequential compositional
structure endowed by the factorization in equation~(\ref{eq:seq-factorization}).
As the particular factorization is a byproduct of the underlying compositional
structure, we can re-write (\ref{eq:seq-factorization}) as a probability
density conditioned on this sequential tree $t$ as follows:

\begin{equation}
p(w_{1}^{n}|t)=\prod_{i=1}^{n}p(w_{1}^{i}|w_{1}^{i-1},w_{i})p(w_{i}).
\end{equation}
Having shown the sequential compositional structure assumption of
the n-gram model, in the next section we try to remove this conditional
density assumption by modeling the joint probability of the sentences
and the compositional trees.

\section{The Compositional Language Model\label{sec:Compostional-Language-Model}}

In this section, we build the framework to carry out the marginalization
over all possible trees. Let $W$ be the sentence and $\mathcal{T}(W)$
be the set of all compositional trees for sentence $W$. The probability
of the sentence $p(W)$ can then be written in terms of the joint
probability over the sentence and compositional structure as
\begin{equation}
p(W)=\sum_{t\in\mathcal{T}(W)}p(W|t)p(t)\label{eq:marginal_1}
\end{equation}

Examining (\ref{eq:marginal_1}), we see that we have two problems
to solve: i) enumerating and building probability distributions over
trees $p(t)$ and ii) modeling the probability of sentences conditioned
on compositional trees $p(W|t)$.

Probabilistic Context Free Grammars (PCFGs) fit the first use case
perfectly. We define a PCFG as a quintuple:

\begin{equation}
(G,\theta)=(N,T,R,S,P)
\end{equation}
where $N$ is the set of non-terminal symbols, $T$ is the set of
terminal symbols, $R$ is the finite set of production rules, $S$
is the special start symbol which is always at the root of a parse
tree and $P$ is the set of probabilities on production rules.\footnote{We restrict our grammar to Chomsky Normal Form (CNF) to simplify the
derivation and explanation. } $\theta$ is a real valued vector of length $|R|$ with the $r$th
index mapping to rule $r\in R$ and value $\theta_{r}\in P$.

Using this definition, we can write the probability of any tree $t$,
$p(t)$, as the product of the production rules used to derive the
sentence $W$ from $S$ i.e.

\begin{equation}
p(t)=\prod_{r\in\tilde{R_{t}}(W)}\theta_{r}\label{eq:pcfg_t_def}
\end{equation}
where $\tilde{R_{t}}(W)\in R$ is the set of production rules used
to derive tree $t$. As we are only interested in the compositional
structure, $\tilde{R}_{t}(W)$ only contains the binary rules for
tree $t$.

\subsection{The Composition Tree Representation\label{sub:Composition-Tree-Representation}}

\begin{figure}[t]
\centering{}\includegraphics[clip,scale=0.8]{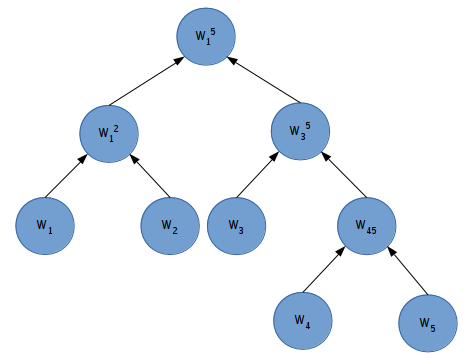}\caption{Parse tree for sentence $W_{12345}$\label{fig:Parse-tree-fig}}
\end{figure}

We now focus our attention to the problem of modeling the sentence
probability conditioned on the compositional tree i.e. $p(W|t)$.
Let $t$ be the compositional tree shown in Figure~\ref{fig:Parse-tree-fig}.
The factorization of $w_{1}^{5}$ given $t$ is

\begin{flalign}
p(w_{1}^{5}|t)= & p(w_{1}^{5}|w_{1}^{2},w_{3}^{5})p(w_{1}^{2}|w_{1},w_{2})\nonumber \\
 & p(w_{3}^{5}|w_{3}w_{4}^{5})p(w_{4}^{5}|w_{4},w_{5})\nonumber \\
 & p(w_{1})p(w_{2})p(w_{3})p(w_{4})p(w_{5}).\label{eq:parse-tree-eq}
\end{flalign}
We now seek to represent any arbitrary tree $t$ such that it can
be factorized easily as we did in (\ref{eq:parse-tree-eq}). We do
this by representing the compositional tree $t$ for a sentence $W$
as a set of compositional rules and leaf nodes. Let us call this set
$R_{t}(W)$. Using this abstraction, the rule set for the sentence
$w_{1}^{5}$ with compositional tree $t$, $R_{t}(w_{1}^{5})$ is

\begin{alignat}{1}
R_{t}(w_{1}^{5})= & \big\{ w_{1}^{5}\leftarrow w_{1}^{2}\;w_{3}^{5},\nonumber \\
 & w_{3}^{5}\leftarrow w_{3}\;w_{4}^{5},\nonumber \\
 & w_{3}w_{4}^{5}\leftarrow w_{4\;}w_{5},w_{4}\nonumber \\
 & w_{1}^{2}\leftarrow w_{1}\;w_{2},w_{1},w_{2}\big\}.\label{eq:rule-set}
\end{alignat}
We now rewrite the factorization in (\ref{eq:parse-tree-eq}) as

\begin{equation}
p(w_{1}^{5}|t)=\prod_{r\in R_{t}(w_{1}^{5})}p(r)
\end{equation}
where $p(pa\leftarrow c_{1}\;c_{2})=p(pa|c_{1},c_{2})$.

In more general terms, let $t$ be the compositional tree for a sentence
$W$. We can write the conditional probability $p(W|t)$ as

\begin{equation}
p(W|t)=\prod_{t\in R_{t}(W)}p(r).\label{eq:p(w|t)_general}
\end{equation}

\subsection{Computing the Sentence Probability\label{sub:Computing-Sentence-Probability}}

Using the definition of $p(t)$ from (\ref{eq:pcfg_t_def}) and the
definition of $p(W|t)$ from (\ref{eq:p(w|t)_general}), we rewrite
the joint probability $p(W,t)$ as
\begin{equation}
p(W,t)=\prod_{c\in R_{t}(W)}p(c)\prod_{r\in\tilde{R}_{t}(W)}\theta_{r}.\label{eq:p(W,t)_subs}
\end{equation}
Now, as the compositional tree $t$ is the same for both the production
rule set $\tilde{R_{t}}(W)$ and the compositional rule set $R_{t}(W)$,
there is a one-to-one mapping between the binary rules in both the
sets.

Adding corresponding POS tags to phrase $w_{i}^{j}$, we can merge
both these sets. Let $R_{t}(W)$ be the merged set. The compositional
rules in this new set can be re-written as 
\begin{equation}
A,w_{i}^{j}\rightarrow BC,w_{i}^{k}w_{k+1}^{j}.
\end{equation}
 Using this new rule set, we can rewrite $p(W,t)$ from (\ref{eq:p(W,t)_subs})
as

\begin{equation}
p(W,t)=\prod_{r\in R_{t}(W)}\zeta_{r}\label{eq:p(W,t)_simpl_in_zeta}
\end{equation}
and $p(W)$ from (\ref{eq:marginal_1}) as
\begin{equation}
p(W)=\sum_{t\in\mathcal{T}(W)}\prod_{r\in R_{t}(W)}\zeta_{r}\label{eq:p(W)_simpl}
\end{equation}
where 
\begin{equation}
\zeta_{r}=\begin{cases}
p(r)\theta_{r} & \mathrm{binary\,rules}\\
p(r) & \mathrm{unary\,rules}
\end{cases}.\label{eq:zeta_r_def}
\end{equation}
 The marginalization formulation in (\ref{eq:p(W)_simpl}) is similar
to one solved by the Inside algorithm.

\textbf{Inside Algorithm:} Let $\pi(A,w_{i}^{j})$ be the inside probability
of $A\in N$ spanning $w_{i}^{j}$. Using this definition, we can
rewrite $p(W)$ in terms of the inside probability as 
\begin{equation}
p(W)=\pi(S,w_{1}^{n}).
\end{equation}
We can now recursively build $\pi(S,w_{1}^{n})$ using a dynamic programming
(DP) based Inside algorithm in the following way:

\textbf{\emph{Base Case}}\textbf{:} For the unary rule, the inside
probability $\pi(A,w_{i})$ is the same as the production rule probability
$\zeta_{A\rightarrow w_{i}}$.

\subparagraph{}

\textbf{\emph{Recursive Definition}}\textbf{:} Let $\pi(B,w_{i}^{k})$
and $\pi(C,w_{k+1}^{j})$ be the inside probabilities spanning $w_{i}^{k}$
rooted at $B$ and $w_{k+1}^{j}$ rooted at $C$ respectively. Let
$r=A,w_{i}^{j}\rightarrow BC,w_{i}^{k}w_{k+1}^{j}$ be the rule which
composes $w_{i}^{j}$ from $w_{i}^{k}$ and $w_{k+1}^{j}$, each one
rooted at $A$, $B$ and $C$ respectively.

The inside probability of rule $r$, $\pi(r)$, can then be calculated
as 
\begin{equation}
\pi(r)=\zeta_{r}\pi(B,w_{i}^{k})\pi(C,w_{k+1}^{j}).
\end{equation}
Let $r_{A,i,k,j}$ be the rule rooted at $A$ spanning $w_{i}^{j}$
splitting at $k$ such that $i<k<j$. We can now calculate $\pi(A,w_{i}^{j})$
by summing over all possible splits between $i,j$. 
\begin{equation}
\pi(A,w_{i}^{j})=\sum_{i\le k<j}\sum_{r_{A,i,k,j}\in R}\pi(r_{A,i,k,j}).
\end{equation}
Examining equations~(\ref{eq:p(W)_simpl}) and (\ref{eq:zeta_r_def}),
we see that we have reduced the problem of modeling $p(W)$ to modeling
$p(r),$ i.e. modeling the probability of compositional rules and
leaf nodes. In the next section we carefully examine this problem.
The approach we follow here is similar to the one taken in \cite{socher2010learning}.
We embed the words in a vocabulary $V$ in a latent space of dimension
$d$, use a compositional function $f$ to build phrases in this latent
space and then build a probability distribution function $p$ over
the leaf nodes and compositional rules.

\subsection{Modeling the compositional probability \label{sub:Modeling-compo-probability}}

The input to our model is an array of integers, each referring to
the index of the word $w$ in our vocabulary $V$. As a first step,
we project each word into a continuous space using $X$, a $d\times|V|$
embedding matrix, to obtain a continuous space vector $x_{i}=X[i]$
corresponding to word $w_{i}$.

A non terminal parent node $pa$ is composed of children nodes $c_{1}$
and $c_{2}$ as
\begin{equation}
pa=f\left(W\left[\begin{array}{c}
c_{1}\\
c_{2}
\end{array}\right]\right)\label{eq:comp}
\end{equation}
where $W$ is a parameter with dimensions $d\times2d$ and $f$ is
a non-linear function like \emph{tanh} or a \emph{sigmoid}. The probability
distribution $p(r)$ over rule $r\in R_{t}(W)$ is modeled as a Gibbs
distribution
\begin{equation}
p(r)=\frac{1}{Z}\exp\left\{ -E(r)\right\} \label{eq:p(r)_def}
\end{equation}
where $E(r)$ is the energy for a compositional rule or a leaf node,
and is modeled as 

\begin{equation}
E(r)=g(u^{T}pa).\label{eq:energy_def}
\end{equation}
Here $u$ is a scoring vector of dimension $d\times1$ and $g$ is
the \emph{identity} function. From (\ref{eq:comp}), (\ref{eq:p(r)_def})
and (\ref{eq:energy_def}), the parameters $\alpha$ of $p(r;\alpha)$
are $(u,X,W)$. 

In the next section we derive an approach for parameter estimation.
We achieve this by formulating training as a maximum likelihood estimation
problem and estimating $\alpha$ by maximizing the probability over
the training set $D$.

\section{Training\label{sec:Training}}

Let $D$ be the set of training sentences. We can write the likelihood
function as
\begin{equation}
\mathcal{L}(\alpha;D)=\prod_{W_{d}\in D}p(W_{d};\alpha).
\end{equation}
This leads to the negative log-likelihood objective function

\begin{equation}
\mathcal{E}_{\mathrm{ML}}(\alpha;D)=-\sum_{W_{d}\in D}\ln(p(W_{d};\alpha)).\label{eq:NLL}
\end{equation}
Substituting the definition of $p(W;\alpha)$ from (\ref{eq:marginal_1})
and $p(W,t;\alpha)$ from (\ref{eq:p(W,t)_simpl_in_zeta}) in (\ref{eq:NLL}),
we get

\begin{equation}
\mathcal{E}_{\mathrm{ML}}(\alpha;D)=-\sum_{W_{d}\in D}\ln\left(\sum_{t\in\mathcal{T}(W_{d})}\prod_{r\in R_{t}(W_{d})}\zeta_{r}(\alpha)\right)\label{eq:NLL_expanded}
\end{equation}
The formulation in equation~(\ref{eq:NLL_expanded}) is very similar
to the standard expectation-maximization (EM) formulation where the
compositional tree $t$ can be seen as a latent variable.

\subsection{Expectation Step}

In the E-step, we compute the expected log-likelihood $\mathcal{Q}(\alpha;\alpha^{old},W)$
as follows.

\begin{multline}
\mathcal{Q}(\alpha;\alpha^{old},W)=\\
-\sum_{t\in\mathcal{T}_{G}(W)}p(t|W;\alpha^{old})\ln(p(t,W;\alpha)).\label{eq:Q_a_a}
\end{multline}
Substituting $p(t,W)$ from (\ref{eq:p(W,t)_simpl_in_zeta}) in (\ref{eq:Q_a_a}),
we can re-write $\mathcal{Q}(\alpha;W)$\footnote{Henceforth we drop the term $\alpha^{old}$ in all our equations for
the sake of brevity.} as

\begin{multline}
\mathcal{Q}(\alpha;W)=\\
-\sum_{t\in\mathcal{T}_{G}(W)}p(t|W)\sum_{r\in R_{t}(W)}\ln(\zeta_{r}(\alpha)).\label{eq:Q_simplified}
\end{multline}
We can simplify the expression further by taking summations over the
trees inside leading to the following expression
\begin{equation}
\mathcal{Q}(\alpha;W)=-\frac{1}{p(W)}\sum_{r\in R}\mu(r)\ln(\zeta_{r}(\alpha))
\end{equation}
where 
\begin{equation}
\mu(r)=\sum_{t\in\mathcal{T}_{G}(W):r\in R_{t}(W)}p(t,W).
\end{equation}
The term $\mu(r)$ sums over all trees that contain rule $r$ and
can be calculated using the inside term $\pi$ and a new term---the
outside term $\beta$. Before computing $\mu(r)$, let's examine how
to compute this outside term $\beta$.

\textbf{Outside Algorithm:} The Inside term $\pi(A,w_{i}^{j})$ is
the probability of $A\in N$ spanning sub-sequence $w_{i}^{j}$. The
Outside term $\beta(A,w_{i}^{j})$ is just the opposite. $\beta(A,w_{i}^{j})$
is the probability of expanding $S$ to sentence $w_{i}^{n}$ such
that sub-sequence $w_{i}^{j}$ rooted at $A$ is left unexpanded.
Similar to the inside probability, the outside probability can be
calculated recursively as follows:

\textbf{\emph{Base Case:}} As the complete sentence is always rooted
at $S$, $\beta(S,w_{1}^{n})$ is always $1$. Moreover, as no other
non-terminal $A$ can be the root of the parse tree, $\beta(A,w_{1}^{n})$,
$A\neq S$ is zero.

\begin{figure}[t]
\begin{centering}
\includegraphics[clip]{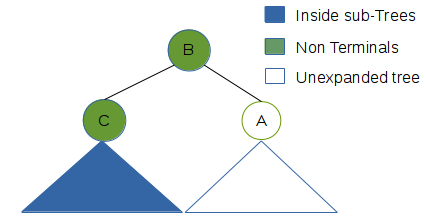}
\par\end{centering}

\caption{Calculating $\beta(B,w_{k}^{j}\rightarrow CA,w_{k}^{i-1}w_{i}^{j})$,
the outside probability of non terminal $A$ spanning $w_{i}^{j}$
such that rule $B,w_{k}^{j}\rightarrow C\,A,w_{k}^{i-1}\,w_{i}^{j}$
was used expand the subsequence to its left\label{fig:B_B_CA_k_i_j}}
\end{figure}

\textbf{\emph{Recursive Definition:}} To compute $\beta(A,w_{i}^{j})$
we need to sum up the probabilities both to the left and right of
$w_{i}^{j}$. Let's consider summing over the left side span $w_{1}^{i-1}$.
Figure~\ref{fig:B_B_CA_k_i_j} shows one of the intermediate steps.
Let $\beta(r_{L})$ be the probability of expanding subsequence $w_{k}^{i-1}$
rooted at $B$ using rule $r_{L}=B,w_{k}^{j}\rightarrow C\,A,w_{k}^{i-1}\,w_{i}^{j}$
such that $w_{i}^{j}$ rooted at $A$ is left unexpanded. 

We can write $\beta(r_{L})$ in terms of its parent's outside probability
$\beta(B,w_{k}^{j})$, left sibling's inside probability $\pi(C,w_{k}^{i-1})$
and $\zeta_{r_{L}}$ as 
\begin{equation}
\beta(r_{L})=\zeta_{r_{L}}\pi(C,w_{k}^{i-1})\beta(B,w_{k}^{j}).\label{eq:beta_left}
\end{equation}
Similarly, on the right hard side of $w_{i}^{j}$, we can calculate
$\beta(r_{R})$ in terms of its parent's outside probability $\beta(B,w_{i}^{k})$,
rule $r_{R}=B,w_{i}^{k}\rightarrow AC,w_{i}^{j}w_{j+1}^{k}$ probability
$\zeta_{r_{R}}$ and inside probability of right sibling $\pi(C,w_{j+1}^{k}$)
as

\begin{equation}
\beta(r_{R})=\zeta_{r_{R}}\pi(C,w_{j+1}^{k})\beta(B,w_{i}^{k}).\label{eq:beta_right}
\end{equation}
Let $r_{A,k,i,j}$ and $r_{A,i,j,k}$ be the rules spanning $w_{k}^{j}$
and $w_{i}^{k}$ such that $A$ spanning $w_{i}^{j}$ is its left
and right child respectively. $\beta(i,j,A)$ can then be calculated
by summing $r_{A,k,i,j}$ and $r_{A,i,j,k}$ over all such rules and
splits, i.e.

\begin{multline}
\beta(A,w_{i}^{j})=\sum_{k=1}^{i-1}\sum_{r_{A,k,i,j}\in R}\beta(r_{A,k,i,j})\\
+\sum_{k=j+1}^{n}\sum_{r_{A,i,j,k}\in R}\beta(r_{A,i,j,k}).
\end{multline}
Now, let's look at the definition of $\mu(r_{A,i,k,j})$. Let $r_{A,i,k,j}$
be a rule rooted at $A$ and span $w_{i}^{j}$. $\pi(r_{A,i,k,j})$
contains probabilities of all the trees that uses production rule
$r_{A,i,k,j}$ in their derivation. $\beta(A,w_{i}^{j})$ would contain
the probability of everything except for $A$ spanning $w_{i}^{j}$.
Hence, their product contains the probability of all parse trees that
have rule $r_{A,i,j}$ in them, i.e. $\mu(r_{A,i,j})$.

\subsection{Minimization Step}

In the M step, the objective is to minimize $\mathcal{Q}(\alpha;\alpha^{old})$
in order to estimate $\alpha^{\ast}$ such that
\begin{multline}
\alpha^{\star}=\arg\min_{\alpha}-\frac{\sum_{r\in R}\ln(\zeta_{r}(\alpha))\mu(r)}{P(W)}.
\end{multline}
Substituting the definition of $\zeta_{r}(\alpha)$ and the value
of $p(r;\alpha)$ from equation~(\ref{eq:p(r)_def}) and differentiating
$\mathbf{\mathcal{Q}(\alpha;\alpha^{old})}$ w.r.t. to $\alpha$,
we get

\begin{equation}
\frac{\partial\mathcal{Q}}{\partial\alpha}=\frac{1}{P(W)}\sum_{r\in R}\left\{ \mu(r)\frac{\partial E(r;\alpha)}{\partial\alpha}\right\} .\label{eq:44}
\end{equation}
Using the definition of the energy function from (\ref{eq:energy_def}),
the partials $\frac{\partial E}{\partial u}$, $\frac{\partial E}{\partial W}$
and $\frac{\partial E}{\partial X}$ are

\begin{equation}
\frac{\partial E(r;u,W,X)}{\partial u}=g^{\prime}(u^{T}pa)pa,
\end{equation}
\begin{equation}
\frac{\partial E(r;u,W,X)}{\partial W}=g^{\prime}(u^{T}pa)\frac{\partial pa}{\partial W},
\end{equation}
and
\begin{equation}
\frac{\partial E(r;u,W,X)}{\partial X}=g^{\prime}(u^{T}pa)\frac{\partial pa}{\partial X}.
\end{equation}
The derivatives $\frac{\partial pa}{\partial W}$ and $\frac{\partial pa}{\partial X}$
can be recursively calculated as follows:

\paragraph{$\mathbf{\partial pa/\partial W}$:}

\textbf{\emph{Base Case:}} For terminal node $p$, $\frac{\partial pa}{\partial W}$
is zero as there is no composition involved.

\textbf{\emph{Recursive Definition:}} Let $\frac{\partial c_{1}}{\partial W}$
and $\frac{\partial c_{2}}{\partial W}$ be the partial derivatives
of children $c_{1}$ and $c_{2}$ respectively. The derivative of
parent embedding $pa$ can then be built using $\frac{\partial c_{1}}{\partial W}$
and $\frac{\partial c_{2}}{\partial W}$ as

\begin{equation}
\frac{\partial pa}{\partial W}=f^{\prime}\times\left\{ \boldsymbol{1}_{j}\circ\bigg[\begin{array}{c}
c_{1}\\
c_{2}
\end{array}\bigg]+W\left[\begin{array}{c}
\frac{\partial c_{1}}{\partial W}\\
\frac{\partial c_{2}}{\partial W}
\end{array}\right]\right\} 
\end{equation}
where $\circ$ is the Hadamard product.

\paragraph{$\mathbf{\partial pa/\partial X}$:}

\textbf{\emph{Base Case:}} Let $x_{i}$ be an embedding vector in
$X$. For a terminal node $p$, there are two possibilities, either
$p$ is equal to $x_{i}$ or it is not. If $p=x_{i}$, $\frac{\partial p}{\partial x_{i}}$
is the identity matrix $I_{d\times d}$ otherwise, it is zero.

\textbf{\emph{Recursive Definition:}} Let $\frac{\partial c_{1}}{\partial x_{i}}$
and $\frac{\partial c_{2}}{\partial x_{i}}$ be the partial derivatives
of children $c_{1}$ and $c_{2}$ respectively w.r.t. $x_{i}$. The
derivative of the parent embedding $pa$ can be built using $\frac{\partial c_{1}}{\partial x_{i}}$
and $\frac{\partial c_{2}}{\partial x_{i}}$ as

\begin{equation}
\frac{\partial pa}{\partial x_{i}}=f^{\prime}\circ W\bigg[\begin{array}{c}
\frac{\partial c_{1}}{\partial x_{i}}\\
\frac{\partial c_{2}}{\partial x_{i}}
\end{array}\bigg].
\end{equation}

\begin{figure}[t]
\begin{centering}
\includegraphics[clip,width=0.5\columnwidth]{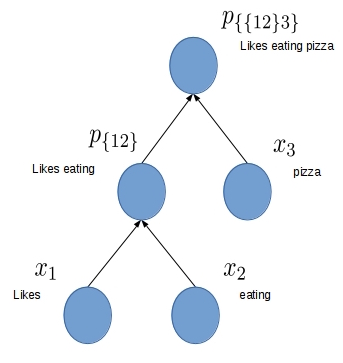}\includegraphics[clip,width=0.5\columnwidth]{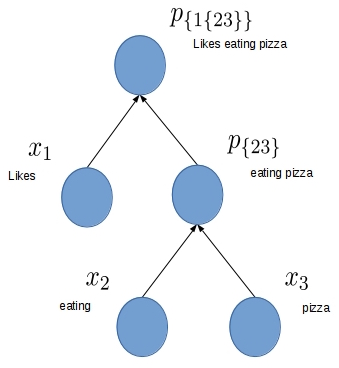}
\par\end{centering}

\caption{Two different compositional trees for $w_{1}^{3}$ leading to different
phrasal representations\emph{.\label{fig:Two-different-composition}}}
\end{figure}

\subsection{Phrasal Representation}

One of the inherent assumptions we made while using the Inside-Outside
Algorithm was that each span has a unique representation. This is
an important assumption because the \emph{states} in the Inside and
Outside algorithms are these spans. A distributed representation breaks
this assumption. To understand this, let's consider a three word sentence
$w_{1}^{3}$. Figure~\ref{fig:Two-different-composition} shows possible
derivations of this sentence. Embeddings for $p_{\{1\{23\}\}}$ and
$p_{\{\{12\}3\}}$ are different as they follow different compositional
paths despite both representing the same phrase $w_{1}^{3}$. Generalizing
this, any sentence or phrase of length $3$ or greater would suffer
from the multiple representation problem due to multiple possible
compositional pathways. 

To understand why this is an issue, let's examine the Inside algorithm
recursion. Dynamic programming works while calculating $\pi(A,w_{i}^{j})$
because we assume that there is only one possible value for each of
$\pi(B,w_{i}^{k})$ and $\pi(C,w_{k+1}^{j})$. As our compositional
probability $p(r)$ depends upon the phrase embeddings, so multiple
possible phrase representations would mean multiple values for $\zeta_{r}$
leading to multiple inside probabilities for each span. This can also
be seen as breakage of the independence assumption of CFGs, as now,
the probability of the parent node also depends upon how it's children
were composed. 

We restore the assumption by taking the expected value of phrasal
embeddings w.r.t. its compositional inside probability $\pi(w_{i}^{j}\rightarrow w_{i}^{k}w_{k+1}^{j})$
\footnote{Inside probability of composition $\pi(w_{i}^{j}\rightarrow w_{i}^{k}w_{k+1}^{j})$
is Inside rule probability $\pi(A,w_{i}^{j}\rightarrow BC,w_{i}^{k}w_{k+1}^{j}$)
marginalized for all non terminals}
\begin{equation}
X(i,j)=E_{\pi}[X(i,k,j)].
\end{equation}
With this approximation, the phrasal embedding for $w_{1}^{3}$ from
Figure~\ref{fig:Two-different-composition} is 
\begin{multline}
X(1,3)=p_{\{1\{23\}\}}\pi(w_{1}^{3}\rightarrow w_{1}w_{2}^{3})+\\
p_{\{\{12\}3\}}\pi(w_{1}^{3}\rightarrow w_{1}^{2}w_{3}).
\end{multline}

This representation is intuitive as well. If composition structures
for a phrase leads to multiple representations, then the best way
to represent that phrase would be an average representation weighted
by the probability of each composition. Now, with the context free
assumption restored, we can use the Inside-Outside algorithm for efficient
marginalization and training.

In the above three sections, we have highlighted the inherent sequential
tree assumption of the traditional models, proposed a compositional
model and derived efficient algorithms to compute $p(W)$ and to train
the parameters. In the next section, we look at how to evaluate our
compositional model. For this we use a  recently proposed discriminative
metric Contrastive Entropy \cite{arora2016contrastive} which doesn't
not require the explicit computation of the partition function.

\section{Evaluation}

The most commonly used metric for benchmarking language models is
perplexity. Despite its widespread use, it cannot evaluate sentence
level models like ours due to its word level model assumption and
reliance on exact probabilities. A recently proposed discriminative
metric Contrastive Entropy \cite{arora2016contrastive} fits our evaluation
use case perfectly. The goal of this new metric is to evaluate the
ability of the model to discriminate between test sentences and their
distorted version.

The Contrastive Entropy, $H_{C}$, is defined as the difference between
the entropy of the test sentence $W_{n}$, and the entropy of the
distorted version of the test sentences $\hat{W}_{n}$ i.e.
\begin{equation}
H_{C}(D;d)=\frac{1}{N}\sum_{W_{n}\in D}H(\hat{W}_{n};d)-H(W_{n})
\end{equation}
Here $D$ is the test set, $d$ is the measure of distortion and $N$,
the number of words or sentences for word level or sentence level
models respectively. 

As this measure is not scale invariant, we also report Contrastive
Entropy Ratio $H_{CR}$ w.r.t. a baseline distortion level $d_{b}$
i.e.
\begin{equation}
H_{CR}(D;d_{b},d)=\frac{H_{C}(D;d)}{H_{C}(D;d_{b})}.
\end{equation}

\subsection{Results}

We use the example dataset provided with the RNNLM toolkit \cite{mikolov2011rnnlm}
for evaluation purposes. The dataset is split into training, testing
and validation set of sizes 10000, 1000 and 1000 sentences respectively.
The training set contains 3720 different words and the test set contains
206 vocabulary words. All reported values here are averaged over 10
runs. 

\begin{figure}[t]
\centering{}\includegraphics[clip,width=1\columnwidth]{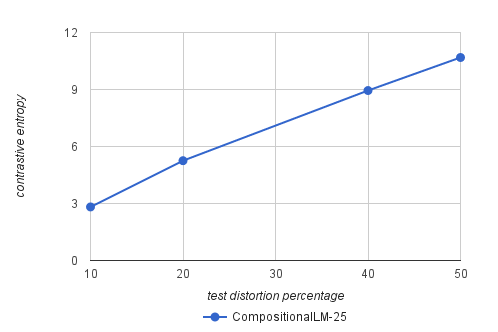}\caption{Contrastive entropy vs distortion levels \label{fig:distortion_vs_cppl}}
\end{figure}

Figure~\ref{fig:distortion_vs_cppl} shows the monotonic increase
in contrastive perplexity as the test distortion level increases.
This is in line with hypothesis that the discriminative ability of
a language model should increase with the test set distortion levels. 

\begin{table}[t]
\begin{centering}
\begin{tabular*}{0.9\columnwidth}{@{\extracolsep{\fill}}>{\centering}p{0.25\columnwidth}>{\centering}p{0.1\columnwidth}>{\centering}p{0.1\columnwidth}>{\centering}p{0.1\columnwidth}>{\centering}p{0.1\columnwidth}}
\toprule 
Model & ppl & $10\%$  & $20\%$  & $40\%$ \tabularnewline
\midrule
3-gram KN & 67.042 & 1.111 & 1.853 & 2.659\tabularnewline
5-gram KN & 66.641 & 1.107 & 1.846 & 2.652\tabularnewline
RNN & 65.361 & 1.322 & 2.231 & 3.227\tabularnewline
cLM-25 & - & 2.818 & 5.252 & 8.945\tabularnewline
cLM-50 & - & 2.833 & 5.441 & 9.179\tabularnewline
\bottomrule
\end{tabular*}
\par\end{centering}

\centering{}\caption{Contrastive entropy at distortion level 10\%, 20\% and 40\%.\label{tab:H_C}}
\end{table}

\begin{table}[t]
\begin{centering}
\begin{tabular*}{0.9\columnwidth}{@{\extracolsep{\fill}}>{\centering}p{0.3\columnwidth}>{\centering}p{0.15\columnwidth}>{\centering}p{0.15\columnwidth}}
\toprule 
Model & $20\%/10\%$  & $40\%/10\%$ \tabularnewline
\midrule
3-gram KN & 1.668 & 2.393\tabularnewline
5-gram KN & 1.667 & 2.395\tabularnewline
RNN & 1.688 & 2.441\tabularnewline
cLM-25 & 1.864 & 3.174\tabularnewline
cLM-50 & 1.921 & 3.240\tabularnewline
\bottomrule
\end{tabular*}
\par\end{centering}

\centering{}\caption{Contrastive entropy ratio at 20\% and 40\% distortion with baseline
distortion of 10\%.\label{tab:H_CR}}
\end{table}

Table~\ref{tab:H_C} compares our language model to standard language
modeling techniques. The n-gram language models here use Kneser Ney
(KN5) smoothing and were generated and evaluated using the SRILM toolkit
\cite{stolcke2002srilm}. The recurrent neural network language model
(RNN)  has a hidden layer of size 400 and was generated using the
RNNLM toolkit. The compositional language models (CLM) in Table~\ref{tab:H_C}
have latent space size of 25 and 50 and were trained using Adagrad
\cite{duchi2011adaptive} with an initial learning rate of 1 and $\ell_{2}$
regularization coefficient of 0.1. CLMs show more than 100\% improvement
over RNNLM, the best performing baseline model, across all distortion
levels. Table~\ref{tab:H_CR} confirms that CLM outperforms all the
baseline models on entropy ratio and isn't impacted by scaling issues.

\section{Conclusion}

In this paper we challenged the linear chain assumption of the traditional
language models by building a model that uses the compositional structure
endowed by context free grammars. We formulated it as a marginalization
problem over the joint probability of sentences and structure and
reduced it to one of modeling compositional rule probabilities $p(r)$.
To the best of our knowledge, this is the first model that looks beyond
the linear chain assumption and uses the compositional structure to
model the language. It is important to note that this compositional
framework is much more general and the way this paper models $p(r)$
is only one of many possible ways to do so.

Also, this paper proposed a compositional framework that recursively
embeds phrases in a latent space and then builds a distribution over
it. This provides us with a distributional language representation
framework which, if trained properly, can be used as a base for various
language processing tasks like NER, POS tagging and sentiment analysis.
The assumption here is that most of the heavy lifting will be done
by the representation and a simple classifier should be able to give
good results over the benchmarks. We also hypothesize that phrasal
embeddings generated using this model will be much more robust and
will also exhibit interesting regularities due to the marginalization
over all possible structures.

As the likelihood optimization proposed here is highly non linear,
better initialization, and improved optimization and regularization
techniques being developed for deep architectures can further improve
these results. Another area of research is to study the effects of
the choice of compositional functions and additional constraints on
representation generated by the model and finally the performance
of the classification layer built on top.\clearpage{}

\bibliographystyle{acl2016_latest}
\bibliography{NLU}

\begin{thebibliography}{}

\bibitem[\protect\citename{Arora and Rangarajan}2016]{arora2016contrastive}
Kushal Arora and Anand Rangarajan.
\newblock 2016.
\newblock Contrastive entropy: A new evaluation metric for unnormalized
  language models.
\newblock {\em arXiv preprint arXiv:1601.00248}.

\bibitem[\protect\citename{Baker and McCallum}1998]{baker1998distributional}
L~Douglas Baker and Andrew~Kachites McCallum.
\newblock 1998.
\newblock Distributional clustering of words for text classification.
\newblock In {\em Proceedings of the 21st annual international ACM SIGIR
  conference on Research and development in information retrieval}, pages
  96--103. ACM.

\bibitem[\protect\citename{Bengio \bgroup et al.\egroup
  }2006]{bengio2006neural}
Yoshua Bengio, Holger Schwenk, Jean-S{\'e}bastien Sen{\'e}cal, Fr{\'e}deric
  Morin, and Jean-Luc Gauvain.
\newblock 2006.
\newblock Neural probabilistic language models.
\newblock In {\em Innovations in Machine Learning}, pages 137--186. Springer.

\bibitem[\protect\citename{Brown \bgroup et al.\egroup }1992]{brown1992class}
Peter~F Brown, Peter~V Desouza, Robert~L Mercer, Vincent J~Della Pietra, and
  Jenifer~C Lai.
\newblock 1992.
\newblock Class-based n-gram models of natural language.
\newblock {\em Computational linguistics}, 18(4):467--479.

\bibitem[\protect\citename{Charniak}2001]{charniak2001immediate}
Eugene Charniak.
\newblock 2001.
\newblock Immediate-head parsing for language models.
\newblock In {\em Proceedings of the 39th Annual Meeting on Association for
  Computational Linguistics}, pages 124--131. Association for Computational
  Linguistics.

\bibitem[\protect\citename{Chelba and Jelinek}2000]{chelba2000structured}
Ciprian Chelba and Frederick Jelinek.
\newblock 2000.
\newblock Structured language modeling.
\newblock {\em Computer Speech \& Language}, 14(4):283--332.

\bibitem[\protect\citename{Chelba \bgroup et al.\egroup
  }1997]{chelba1997structure}
Ciprian Chelba, David Engle, Frederick Jelinek, Victor Jimenez, Sanjeev
  Khudanpur, Lidia Mangu, Harry Printz, Eric Ristad, Ronald Rosenfeld, Andreas
  Stolcke, et~al.
\newblock 1997.
\newblock Structure and performance of a dependency language model.
\newblock In {\em EUROSPEECH}. Citeseer.

\bibitem[\protect\citename{Duchi \bgroup et al.\egroup
  }2011]{duchi2011adaptive}
John Duchi, Elad Hazan, and Yoram Singer.
\newblock 2011.
\newblock Adaptive subgradient methods for online learning and stochastic
  optimization.
\newblock {\em The Journal of Machine Learning Research}, 12:2121--2159.

\bibitem[\protect\citename{Goodman}2001]{goodman2001bit}
Joshua~T Goodman.
\newblock 2001.
\newblock A bit of progress in language modeling.
\newblock {\em Computer Speech \& Language}, 15(4):403--434.

\bibitem[\protect\citename{Kuhn and De~Mori}1990]{kuhn1990cache}
Roland Kuhn and Renato De~Mori.
\newblock 1990.
\newblock A cache-based natural language model for speech recognition.
\newblock {\em Pattern Analysis and Machine Intelligence, IEEE Transactions
  on}, 12(6):570--583.

\bibitem[\protect\citename{Lari and Young}1990]{lari1990estimation}
Karim Lari and Steve~J Young.
\newblock 1990.
\newblock The estimation of stochastic context-free grammars using the
  inside-outside algorithm.
\newblock {\em Computer speech \& language}, 4(1):35--56.

\bibitem[\protect\citename{Lau \bgroup et al.\egroup }1993]{lau1993trigger}
Raymond Lau, Ronald Rosenfeld, and Salim Roukos.
\newblock 1993.
\newblock Trigger-based language models: A maximum entropy approach.
\newblock In {\em Acoustics, Speech, and Signal Processing, 1993. ICASSP-93.,
  1993 IEEE International Conference on}, volume~2, pages 45--48. IEEE.

\bibitem[\protect\citename{Mikolov \bgroup et al.\egroup
  }2010]{mikolov2010recurrent}
Tomas Mikolov, Martin Karafi{\'a}t, Lukas Burget, Jan Cernock{\`y}, and Sanjeev
  Khudanpur.
\newblock 2010.
\newblock Recurrent neural network based language model.
\newblock In {\em INTERSPEECH}, pages 1045--1048.

\bibitem[\protect\citename{Mikolov \bgroup et al.\egroup
  }2011a]{mikolov2011extensions}
Tomas Mikolov, Stefan Kombrink, Lukas Burget, JH~Cernocky, and Sanjeev
  Khudanpur.
\newblock 2011a.
\newblock Extensions of recurrent neural network language model.
\newblock In {\em Acoustics, Speech and Signal Processing (ICASSP), 2011 IEEE
  International Conference on}, pages 5528--5531. IEEE.

\bibitem[\protect\citename{Mikolov \bgroup et al.\egroup
  }2011b]{mikolov2011rnnlm}
Tomas Mikolov, Stefan Kombrink, Anoop Deoras, Lukar Burget, and Jan Cernocky.
\newblock 2011b.
\newblock {RNNLM}-recurrent neural network language modeling toolkit.
\newblock In {\em Proc. of the 2011 ASRU Workshop}, pages 196--201.

\bibitem[\protect\citename{Mnih and Hinton}2009]{mnih2009scalable}
Andriy Mnih and Geoffrey~E Hinton.
\newblock 2009.
\newblock A scalable hierarchical distributed language model.
\newblock In {\em Advances in neural information processing systems}, pages
  1081--1088.

\bibitem[\protect\citename{Morin and Bengio}2005]{morin2005hierarchical}
Frederic Morin and Yoshua Bengio.
\newblock 2005.
\newblock Hierarchical probabilistic neural network language model.
\newblock In {\em AISTATS}, volume~5, pages 246--252. Citeseer.

\bibitem[\protect\citename{Pereira \bgroup et al.\egroup
  }1993]{pereira1993distributional}
Fernando Pereira, Naftali Tishby, and Lillian Lee.
\newblock 1993.
\newblock Distributional clustering of english words.
\newblock In {\em Proceedings of the 31st annual meeting on Association for
  Computational Linguistics}, pages 183--190. Association for Computational
  Linguistics.

\bibitem[\protect\citename{Socher \bgroup et al.\egroup
  }2010]{socher2010learning}
Richard Socher, Christopher~D Manning, and Andrew~Y Ng.
\newblock 2010.
\newblock Learning continuous phrase representations and syntactic parsing with
  recursive neural networks.
\newblock In {\em Proceedings of the NIPS-2010 Deep Learning and Unsupervised
  Feature Learning Workshop}, pages 1--9.

\bibitem[\protect\citename{Socher \bgroup et al.\egroup
  }2013]{socher2013parsing}
Richard Socher, John Bauer, Christopher~D Manning, and Andrew~Y Ng.
\newblock 2013.
\newblock Parsing with compositional vector grammars.
\newblock In {\em In Proceedings of the ACL conference}. Citeseer.

\bibitem[\protect\citename{Stolcke}2002]{stolcke2002srilm}
Andreas Stolcke.
\newblock 2002.
\newblock Srilm-an extensible language modeling toolkit.
\newblock In {\em INTERSPEECH}.

\end{thebibliography}

\end{document}